\documentclass[11pt,a4paper]{article}
\pdfoutput=1
\usepackage[hyperref]{naaclhlt2018}
\usepackage{times}
\usepackage{latexsym}

\usepackage{graphicx}
\usepackage{tabularx}
\usepackage{multirow}
\usepackage{marvosym} 

\usepackage{vntex}
\usepackage[english]{babel}

\usepackage{url}

\usepackage{listings}

\aclfinalcopy 

\title{VnCoreNLP: A Vietnamese Natural Language Processing Toolkit}

\author{Thanh Vu$^1$, Dat Quoc Nguyen$^2$, Dai Quoc Nguyen$^3$, Mark Dras$^4$ \and Mark Johnson$^4$\\
$^1$Newcastle University, United Kingdom; $^2$The University of Melbourne, Australia; \\
$^3$Deakin University, Australia; $^4$Macquarie University, Australia \\
{\tt thanh.vu@newcastle.ac.uk}, {\tt dqnguyen@unimelb.edu.au},\\
{\tt dai.nguyen@deakin.edu.au}, {\tt\{mark.dras, mark.johnson\}@mq.edu.au}}

\date{}

\begin{document}
\maketitle
\begin{abstract}
We present an easy-to-use and fast toolkit, namely VnCoreNLP---a Java NLP annotation pipeline for Vietnamese. Our VnCoreNLP supports key natural language processing (NLP) tasks including  word segmentation, part-of-speech (POS) tagging, named entity recognition (NER) and dependency parsing, and obtains state-of-the-art (SOTA) results for these tasks. 
We release VnCoreNLP to provide  rich linguistic annotations to facilitate research work on Vietnamese NLP. 
Our VnCoreNLP is open-source  and available at: \url{https://github.com/vncorenlp/VnCoreNLP}.
\end{abstract}

\section{Introduction}

Research on Vietnamese NLP has been actively explored in the last decade, boosted by the successes of the 4-year KC01.01/2006-2010  national project  on Vietnamese language and speech processing (VLSP). Over the last 5 years, standard benchmark datasets for key Vietnamese NLP tasks are publicly available: datasets for word segmentation and POS tagging were released for the first VLSP evaluation campaign in 2013; a dependency treebank was published in 2014 \cite{Nguyen2014NLDB}; and an NER dataset was released for the second  VLSP  campaign in 2016.  So there is a need for building an NLP pipeline, such as the Stanford CoreNLP toolkit \cite{manning-EtAl:2014:P14-5}, for those key tasks to assist users and to support researchers and tool developers of downstream tasks.

\newcite{NguyenPN2010} and \newcite{Le:2013:VOS} built Vietnamese NLP pipelines by wrapping  existing  word segmenters and POS taggers including: JVnSegmenter \cite{Y06-1028}, vnTokenizer \cite{Le2008}, JVnTagger \cite{NguyenPN2010} and  vnTagger \cite{lehong00526139}. However,
these word segmenters and POS taggers are no longer considered
SOTA  models for Vietnamese \cite{NguyenL2016,JCSCE}.  \newcite{PhamPNP2017b} built the NNVLP toolkit for Vietnamese sequence labeling tasks by applying a  BiLSTM-CNN-CRF model \cite{ma-hovy:2016:P16-1}. However,  \newcite{PhamPNP2017b}  did not make a comparison to SOTA traditional feature-based models. In addition,   NNVLP   is slow with a processing speed at about 300 words per second, which is not practical for real-world application such as dealing  with  large-scale data.

\setlength{\abovecaptionskip}{5pt plus 2pt minus 1pt}
\begin{figure}[!t]
\centering
\includegraphics[width=7.5cm]{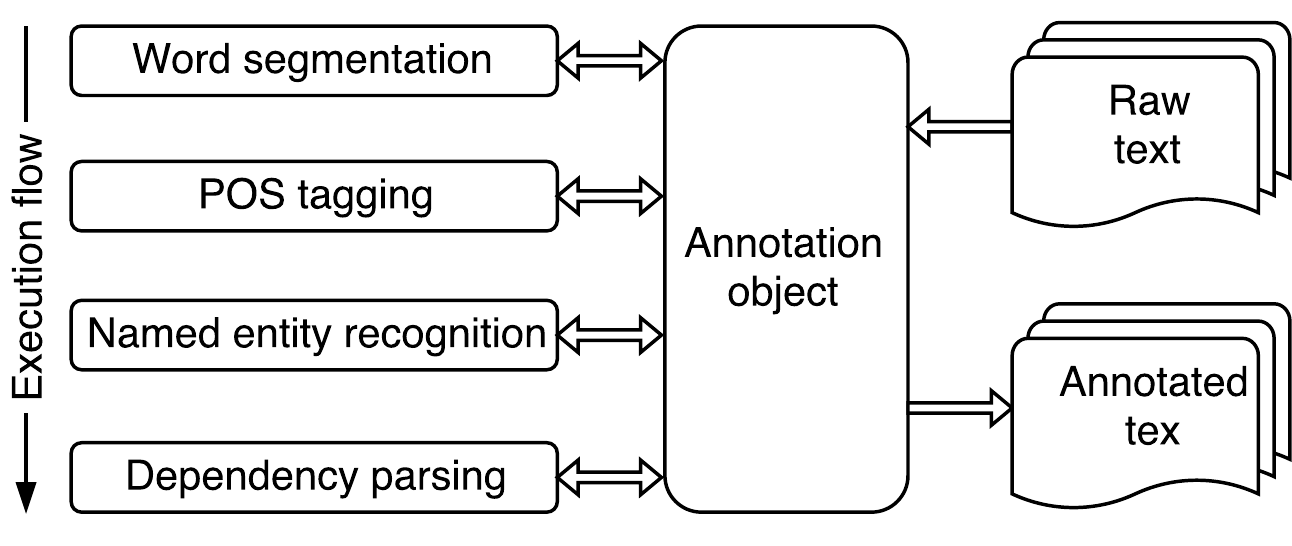} 
\caption{In pipeline architecture of VnCoreNLP, annotations are performed on an {\tt Annotation} object.}
\label{fig:diagram}
\end{figure}

In this paper, we present a Java NLP toolkit for Vietnamese, namely  VnCoreNLP, which aims to facilitate Vietnamese NLP research   by providing rich linguistic annotations through key NLP components of word segmentation, POS tagging, NER and dependency parsing. Figure \ref{fig:diagram} describes the overall system architecture. The
following items highlight typical characteristics of VnCoreNLP:

\begin{itemize}
\setlength{\itemsep}{5pt}
\setlength{\parskip}{0pt}
\setlength{\parsep}{0pt}

\item \textbf{Easy-to-use} -- All VnCoreNLP components are wrapped into a single .jar file, so users do not have to install external dependencies. Users can run processing pipelines  from either the command-line or the Java API.

\item \textbf{Fast} -- VnCoreNLP is fast, so it can be used  for dealing with large-scale data. Also  it benefits users  suffering from limited computation resources (e.g. users from  Vietnam).

\item \textbf{Accurate} -- VnCoreNLP components obtain higher results than all previous published results on  the same benchmark datasets.

\end{itemize}

\section{Basic usages}

Our design goal is to make VnCoreNLP simple to setup and run from either the command-line or the Java API. Performing linguistic annotations for a given file can be done by using a simple command as in Figure  \ref{fig:command}.

\begin{figure}[ht]
{\footnotesize\ttfamily \$ java -Xmx2g -jar VnCoreNLP.jar -fin input.txt -fout output.txt}
\caption{Minimal command to run VnCoreNLP.}
\label{fig:command}
\end{figure}

Suppose that the file {\ttfamily input.txt} in Figure  \ref{fig:command} contains a sentence ``Ông Nguyễn Khắc Chúc đang làm việc tại Đại học Quốc gia Hà Nội.'' (Mr\textsubscript{Ông} Nguyen Khac Chuc  is\textsubscript{đang} working\textsubscript{làm\_việc} at\textsubscript{tại} Vietnam National\textsubscript{quốc\_gia} University\textsubscript{đại\_học} Hanoi\textsubscript{Hà\_Nội}). Table \ref{tab:expoutput} shows the output for this sentence in plain text form.

\begin{table}[ht]
    \centering
    \resizebox{8cm}{!}{
    \begin{tabular}{l l l l l l}
        1 & Ông & Nc & O & 4 & sub \\
        2 & Nguyễn\_Khắc\_Chúc  & Np & B-PER & 1 & nmod\\
        3 & đang & R & O & 4 & adv\\
        4 & làm\_việc & V & O & 0 & root\\
        5 & tại & E & O & 4 & loc\\
        6 & Đại\_học & N & B-ORG & 5 & pob\\
        7 & Quốc\_gia & N & I-ORG & 6 & nmod\\
        8 & Hà\_Nội & Np & I-ORG & 6 & nmod\\
        9 & . & CH & O & 4 & punct\\
    \end{tabular}
    } 
    \caption{The output in file {\ttfamily output.txt} for the sentence `Ông Nguyễn Khắc Chúc đang làm việc tại Đại học Quốc gia Hà Nội.'' from  file {\ttfamily input.txt} in Figure \ref{fig:command}. The output is  in a 6-column format representing word index, word form, POS tag, NER label, head index of the current word, and dependency relation type.}
    \label{tab:expoutput}
\end{table}

 Similarly, we can also get the same output by using the API as easy as in Listing \ref{lst1}.

\begin{lstlisting}[label=lst1,caption= {Minimal code for an analysis pipeline.}]
VnCoreNLP pipeline = new VnCoreNLP() ;
Annotation annotation = new Annotation("%*Ông Nguyễn Khắc Chúc đang làm việc tại Đại học Quốc gia Hà Nội.*)");
pipeline.annotate(annotation);
String annotatedStr = annotation.toString();
\end{lstlisting}

In addition, 
Listing \ref{lst2} provides a more realistic and complete example code, presenting key components of the toolkit. 
Here an annotation pipeline can be used for any text rather than just a single sentence, e.g. for  a paragraph or entire news story.

\section{Components}

This section  briefly describes each component of VnCoreNLP. Note that our goal is not to develop new approach or model for each component task. Here we focus on incorporating existing models into a single pipeline. In particular, except a new model we develop for the language-dependent component of word segmentation, we apply traditional feature-based models which obtain SOTA results for English POS tagging, NER and dependency parsing to Vietnamese.  The reason is based on a well-established belief in the literature that for  a less-resourced language such as Vietnamese, we should consider using  feature-based models to obtain fast and accurate performances, rather than using neural network-based  models \cite{King2015}.

\setlength{\textfloatsep}{1pt plus 1.0pt minus 1.0pt}
\begin{lstlisting}[float=tp,label=lst2,caption= {A simple and complete example code.}]
import vn.pipeline.*;
import java.io.*;
public class VnCoreNLPExample {
 public static void main(String[] args) throws IOException {
  // "wseg", "pos", "ner", and "parse" refer to as word segmentation, POS tagging, NER and dependency parsing, respectively.
  String[] annotators = {"wseg", "pos", "ner", "parse"};
  VnCoreNLP pipeline = new VnCoreNLP(annotators);
  // Mr Nguyen Khac Chuc is working at Vietnam National University, Hanoi. Mrs Lan, Mr Chuc's wife, is also working at this university.
  String str = %*"Ông Nguyễn Khắc Chúc  đang làm việc tại Đại học Quốc gia Hà Nội. Bà Lan, vợ ông Chúc, cũng làm việc tại đây."*);
  Annotation annotation = new Annotation(str);
  pipeline.annotate(annotation);
  PrintStream outputPrinter = new PrintStream("output.txt");
  pipeline.printToFile(annotation, outputPrinter);
  // Users can get a single sentence to analyze individually
  Sentence firstSentence = annotation.getSentences().get(0);
 }
}
\end{lstlisting}
\begin{itemize}
\setlength{\itemsep}{5pt}
\setlength{\parskip}{0pt}
\setlength{\parsep}{0pt}

\item \textbf{wseg} -- Unlike English where white space is a strong indicator of word boundaries, when written in Vietnamese white space  is also used to separate syllables that constitute words. So word segmentation  is referred to as the key first  step in Vietnamese NLP{.}\ {W}e have proposed a  transformation rule-based learning model for Vietnamese word segmentation, which obtains better segmentation accuracy and speed than all previous word segmenters. See details in \newcite{NguyenNVDJ2018}.

\item \textbf{pos} -- To label words with their POS tag, we apply   MarMoT  which is a generic
CRF framework and a SOTA POS and morphological
tagger  \cite{mueller-schmid-schutze:2013:EMNLP}.\footnote{\url{http://cistern.cis.lmu.de/marmot/}}

\item \textbf{ner} -- To recognize named entities, we apply a dynamic feature induction model that automatically optimizes feature combinations \cite{choi:2016:N16-1}.\footnote{\url{https://emorynlp.github.io/nlp4j/components/named-entity-recognition.html}}

\item \textbf{parse} -- To perform dependency parsing, we apply the greedy version of a transition-based parsing model with selectional branching \cite{choi2015ACL}.\footnote{\url{https://emorynlp.github.io/nlp4j/components/dependency-parsing.html}}
\end{itemize}

\section{Evaluation}

We detail experimental results of the  word segmentation (\textbf{wseg}) and POS tagging (\textbf{pos}) components of VnCoreNLP in \newcite{NguyenNVDJ2018} and \newcite{NguyenVNDJ-ALTA-2017}, respectively. In particular, our word segmentation component gets the highest results in terms of both segmentation F1 score at 97.90\% and speed  at 62K words per second.\footnote{All speeds reported in this paper are  computed on a personal computer of Intel Core i7 2.2 GHz.} Our POS tagging component also obtains the highest  accuracy to date at 95.88\% with a fast tagging speed at 25K words per second, and outperforms BiLSTM-CRF-based models. 
Following subsections present evaluations for the NER (\textbf{ner}) and dependency parsing (\textbf{parse}) components.

\subsection{Named entity recognition}\label{ssec:ner}

We make a comparison between SOTA feature-based and neural network-based models, which, to the best of our knowledge, has not been done in any prior work on   Vietnamese NER.

\paragraph{Dataset:} The NER shared task at the 2016 VLSP workshop provides a set of 16,861 manually annotated sentences for training and development, and a set of 2,831 manually annotated sentences for test, with four NER labels PER, LOC, ORG and MISC. Note that in both  datasets, words are also supplied with gold POS tags. In addition, each word representing a full personal name are separated into syllables  that constitute the word. 
So this annotation scheme results in an unrealistic scenario for a pipeline evaluation because:  (\textbf{i}) gold POS tags are not available in a real-world application, and (\textbf{ii}) in the standard annotation (and benchmark datasets) for Vietnamese word segmentation  and POS tagging  \cite{nguyen-EtAl:2009:LAW-III},   each full name is referred to as a word token (i.e.,  all  word segmenters have been trained to output a full name as a word  and all POS taggers have been trained to assign a  label to the entire full-name).  

For a more realistic scenario, we merge those contiguous syllables constituting a full name to form a word.\footnote{Based on the gold label PER,  contiguous syllables such as ``Nguyễn/B-PER'', ``Khắc/I-PER'' and ``Chúc/I-PER'' are merged to form a word  as ``Nguyễn\_Khắc\_Chúc/B-PER.''} Then we replace the gold POS tags by automatic tags predicted by our POS tagging component. From the set of 16,861 sentences, we sample 2,000 sentences for development and using the remaining 14,861 sentences for training.

\paragraph{Models:} We make an empirical comparison between the VnCoreNLP's NER component  and the following neural network-based models:
\begin{itemize}
\setlength{\itemsep}{5pt}
\setlength{\parskip}{0pt}
\setlength{\parsep}{0pt}

    \item {BiLSTM-CRF} \cite{HuangXY15}  is a sequence labeling model which extends the BiLSTM model with a CRF layer.

    \item  {BiLSTM-CRF + CNN-char}, i.e. {BiLSTM-CNN-CRF}, is an extension of  {BiLSTM-CRF}, using CNN to derive character-based word representations   \cite{ma-hovy:2016:P16-1}.
    
    \item {BiLSTM-CRF + LSTM-char}   is an extension of  {BiLSTM-CRF}, using BiLSTM to derive the character-based word representations \cite{lample-EtAl:2016:N16-1}.
    
    \item BiLSTM-CRF\textsubscript{+POS} is another extension to BiLSTM-CRF,   incorporating embeddings of automatically predicted POS tags \cite{reimers-gurevych:2017:EMNLP2017}.

\end{itemize}

We use a well-known implementation which is optimized for performance of all  BiLSTM-CRF-based models from \newcite{reimers-gurevych:2017:EMNLP2017}.\footnote{\url{https://github.com/UKPLab/emnlp2017-bilstm-cnn-crf}} We then follow \newcite[Section 3.4]{NguyenVNDJ-ALTA-2017} to perform hyper-parameter tuning.\footnote{We employ pre-trained Vietnamese word vectors from \url{https://github.com/sonvx/word2vecVN}.}

\setlength{\textfloatsep}{20.0pt plus 2.0pt minus 4.0pt}
\begin{table}[!t]
    \centering
     \begin{tabular}{l|c|l}
    \hline
     \textbf{Model}  & \textbf{F1} & \textbf{Speed}  \\
     \hline
     VnCoreNLP &  \textbf{88.55}  &  \textbf{18K} \\
      BiLSTM-CRF & 86.48 & 2.8K   \\
      \ \ \ \ \ + CNN-char & {88.28} &  1.8K  \\
       \ \ \ \ \ + LSTM-char & 87.71 &  1.3K   \\
      BiLSTM-CRF\textsubscript{+POS} & 86.12  & \_    \\
      \ \ \ \ \ + CNN-char & 88.06 & \_      \\
      \ \ \ \ \ + LSTM-char & 87.43 & \_   \\
    \hline
    \end{tabular}
    \caption{F1 scores (in \%) on the test set  w.r.t. gold word-segmentation. ``\textbf{Speed}'' denotes the processing speed of the number of words per second (for VnCoreNLP, we include the time POS tagging takes in the speed).}
    \label{tab:ner}
\end{table}

\paragraph{Main results:} Table \ref{tab:ner} presents F1 score  and  speed of each model on the test set, where VnCoreNLP obtains the highest  score at 88.55\% with a fast speed at 18K words per second. In particular, VnCoreNLP  obtains  10 times faster speed than  the second most accurate model BiLSTM-CRF + CNN-char. 

It is initially surprising that for such an  isolated language as Vietnamese  where all words are not inflected, using character-based representations  helps producing 1+\% improvements to the BiLSTM-CRF model. We  find that the improvements to BiLSTM-CRF are mostly accounted for by the PER label. The reason turns out to be simple: about 50\% of named entities are labeled with tag PER, so  character-based representations are in fact able to capture  common family, middle or given name syllables in `unknown' full-name  words.  Furthermore, we also find that BiLSTM-CRF-based models do not benefit from  additional predicted POS tags. It is probably  because BiLSTM can take word order into account, while without word inflection,  all grammatical information in Vietnamese is conveyed through its fixed word order, thus explicit predicted POS tags with noisy grammatical information are not helpful.

\subsection{Dependency parsing}\label{ssec:dep}

\paragraph{Experimental setup:} We use the Vietnamese dependency treebank VnDT \cite{Nguyen2014NLDB} consisting of 10,200 sentences in our experiments. Following \newcite{NguyenALTA2016}, we use the last 1020 sentences of VnDT for test while the remaining sentences are used for training. Evaluation metrics are the labeled attachment score (LAS) and unlabeled attachment score (UAS).

\paragraph{Main results:} Table \ref{tab:dep} compares the dependency parsing results of VnCoreNLP with results reported in prior work, using the same experimental setup. The first six rows present the scores with gold POS tags. The next two rows show  scores of VnCoreNLP with automatic  POS tags which are produced by our POS tagging component. The last row presents scores of the joint POS tagging and dependency parsing model jPTDP  \protect\cite{NguyenCoNLL2017}. Table \ref{tab:dep} shows that compared to previously published results,  VnCoreNLP produces the highest LAS score. Note that previous results for other systems are reported without using additional information of automatically predicted NER labels. In this  case,  the LAS score  for VnCoreNLP without automatic NER features (i.e. VnCoreNLP\textsubscript{--NER} in Table \ref{tab:dep}) is still higher than previous ones. Notably, we also obtain a fast parsing speed at 8K words per second.

\begin{table}[!t]
    \centering
    \setlength{\tabcolsep}{0.5em}
    \def\arraystretch{1.1}
    \begin{tabular}{l|l|c|c|l }
    \hline
     \multicolumn{2}{c|}{\textbf{Model}}  & \textbf{LAS} & \textbf{UAS} & \textbf{Speed}  \\ 
     \hline
    \multirow{6}{*}{\rotatebox[origin=c]{90}{Gold POS}}  
    &  VnCoreNLP  &  \textbf{73.39}  &  79.02  & \_ \\
    & VnCoreNLP\textsubscript{--NER}  & 73.21  & 78.91 & \_ \\
    & BIST-bmstparser &  73.17 & \textbf{79.39} & \_ \\
    & BIST-barchybrid & 72.53 & 79.33 & \_ \\
    & MSTParser & 70.29 & 76.47 & \_\\
    & MaltParser & 69.10 & 74.91 & \_\\
    \hline
    \multirow{3}{*}{\rotatebox[origin=c]{90}{Auto POS}}  
      &   VnCoreNLP  & \textbf{70.23}  &  76.93 & 8K \\
      & VnCoreNLP\textsubscript{--NER}  & 70.10 & 76.85 &  \textbf{9K} \\
      & jPTDP & 69.49 & \textbf{77.68} & 700 \\
    \hline
    \end{tabular}
    \caption{LAS and UAS scores (in \%) computed on all tokens (i.e. including punctuation) on the test set w.r.t. gold word-segmentation. ``\textbf{Speed}'' is defined as in Table \ref{tab:ner}. The subscript ``--NER'' denotes the model without using automatically predicted NER labels as features. The results of the MSTParser \protect\cite{McDonald2005OLT}, MaltParser \protect\cite{Nivre2007}, and BiLSTM-based parsing models BIST-bmstparser and BIST-barchybrid \protect\cite{TACL885}  are reported in \protect\newcite{NguyenALTA2016}. The result of the jPTDP model  for Vietnamese is mentioned in \protect\newcite{NguyenVNDJ-ALTA-2017}.}
    \label{tab:dep}
\end{table}

\section{Conclusion}

In this paper, we have presented the VnCoreNLP toolkit---an easy-to-use, fast and accurate   processing pipeline for Vietnamese NLP. VnCoreNLP provides core NLP steps including word segmentation, POS tagging, NER and dependency parsing. Current version of VnCoreNLP has been trained without any linguistic optimization, i.e. we only employ existing pre-defined features in the traditional feature-based models for POS tagging, NER and dependency parsing. So future work will   focus on incorporating Vietnamese linguistic features into these feature-based models.  

VnCoreNLP is released  for research and educational purposes, and available at: \url{https://github.com/vncorenlp/VnCoreNLP}.

\bibliography{Refs}
\bibliographystyle{naacl_natbib}

\end{document}